\documentclass[final]{cvpr}

\usepackage{times}
\usepackage{epsfig}
\usepackage{graphicx}
\usepackage{amsmath}
\usepackage{amssymb}

\usepackage{multirow}
\usepackage{enumitem} 
\usepackage{subcaption} 
\usepackage{amsbsy} 

\graphicspath{ {images/} }

\usepackage[pagebackref=true,breaklinks=true,colorlinks,bookmarks=false]{hyperref}

\def\cvprPaperID{1429}
\def\confYear{CVPR 2021}
\begin{document}


\setlength{\abovecaptionskip}{1pt}

\setlength{\belowcaptionskip}{-10pt}

\title{Monocular Real-time Full Body Capture with Inter-part Correlations}

\author{
  \normalsize{Yuxiao Zhou\textsuperscript{1}}
  \quad
  \normalsize{Marc Habermann\textsuperscript{2,3}}
  \quad
  \normalsize{Ikhsanul Habibie\textsuperscript{2,3}}
  \quad
  \normalsize{Ayush Tewari\textsuperscript{2,3}}
  \quad
  \normalsize{Christian Theobalt\textsuperscript{2,3}}
  \quad
  \normalsize{Feng Xu\textsuperscript{1}\thanks{This work was supported by the National Key R\&D Program of China 2018YFA0704000, the NSFC (No.61822111, 61727808), Beijing Natural Science Foundation (JQ19015), and the ERC Consolidator Grant 4DRepLy (770784). Feng Xu is the corresponding author.}}\\
  \small{\textsuperscript{1}BNRist and School of Software, Tsinghua University}
  \quad
  \small{\textsuperscript{2}Max Planck Institute for Informatics}
  \quad
  \small{\textsuperscript{3}Saarland Informatics Campus}
}

\maketitle

\begin{abstract}
We present the first method for real-time full body capture that estimates shape and motion of body and hands together with a dynamic 3D face model from a single color image.
Our approach uses a new neural network architecture that exploits correlations between body and hands at high computational efficiency.
Unlike previous works, our approach is jointly trained on multiple datasets focusing on hand, body or face separately, without requiring data where all the parts are annotated at the same time, which is much more difficult to create at sufficient variety.
The possibility of such multi-dataset training enables superior generalization ability.
In contrast to earlier monocular full body methods, our approach captures more expressive 3D face geometry and color by estimating the shape, expression, albedo and illumination parameters of a statistical face model.
Our method achieves competitive accuracy on public benchmarks, while being significantly faster and providing more complete face reconstructions.

\end{abstract}


\section{Introduction}
%
Human motion capture from a single color image is an important and widely studied topic in computer vision.
Most solutions are unable to capture local motions of hands and faces together with full body motions.
This renders them unsuitable for a variety of applications, e.g. AR, VR, or tele-presence, where capturing full human body pose and shape, including hands and face, is highly important.
In these applications, monocular approaches should ideally recover the full body pose (including facial expression) as well as a render-ready dense surface which contains person-specific information, such as facial identity and body shape.
Moreover, they should run at real-time framerates.
Much progress has been made on relevant subtasks, i.e. body pose estimation~\cite{kolotouros2019learning,kanazawa2018end,omran2018neural,mehta2017vnect}, hand pose estimation~\cite{zhou2020monocular,mueller2018ganerated,zimmermann2019freihand}, and face capture~\cite{egger20203d,tewari2017mofa,tewari2018self,sengupta2018sfsnet,zollhofer2018state}.
However, joint \textit{full body} capture, let alone in real-time, is still an open problem.
Several recent works~\cite{choutas2020monocular,xiang2019monocular,jin2020whole,pavlakos2019expressive,martinez2019single} have demonstrated promising results on capturing the full body.
Nevertheless, they either only recover sparse 2D keypoints~\cite{martinez2019single,jin2020whole}, require specific training data~\cite{choutas2020monocular,jin2020whole} where body, hands, and face are annotated altogether which is expensive to collect, or cannot achieve real-time performance~\cite{choutas2020monocular,xiang2019monocular,pavlakos2019expressive,martinez2019single}.
%
%
\begin{figure}[t]
  \centering
  \includegraphics[width=\linewidth]{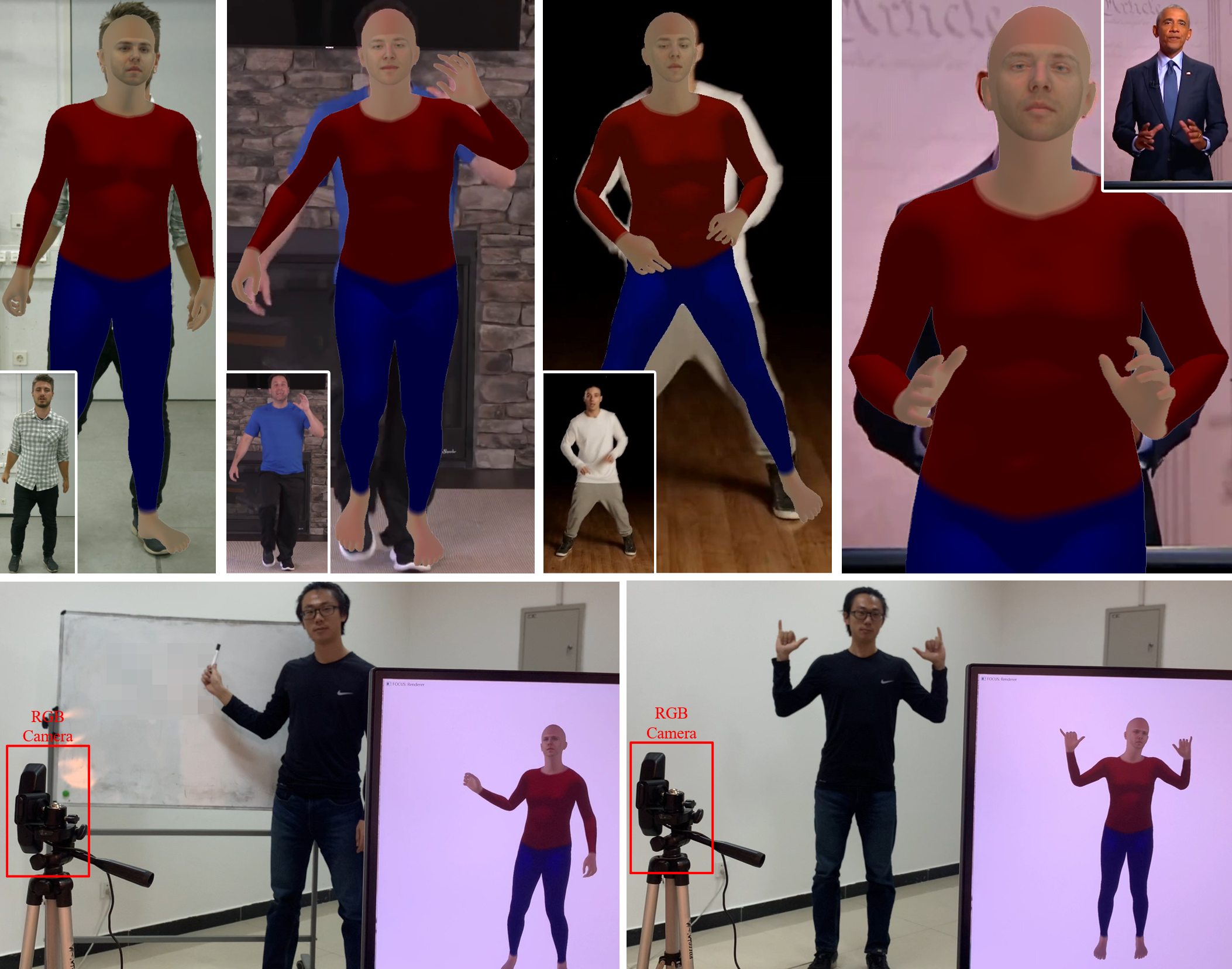}
  \caption{
    We present the first real-time monocular approach that jointly captures shape and pose of body and hands together with facial geometry and color.
    Top: results on in-the-wild sequences.
    Bottom: real-time demo.
    Our approach predicts facial color while the body color is set manually.
  }
  \label{fig:teaser}
\end{figure}
%
\par
%
We therefore introduce the first real-time monocular approach that estimates:
\textit{1)} 2D and 3D keypoint positions of body and hands;
\textit{2)} 3D joint angles and shape parameters of body and hands;
and
\textit{3)} shape, expression, albedo, and illumination parameters of a 3D morphable face model~\cite{tewari2017mofa,egger20203d}.
To recover the dense mesh, we use the SMPLH model~\cite{romero2017embodied} for body and hands surface, and replace its face area with a more expressive face model.
%
\par
%
To achieve real-time performance without the loss of accuracy, we rigorously design our new network architecture to exploit inter-part correlations by streaming body features into the hand pose estimation branch.
Specifically, the sub-network for hand keypoint detection takes in two sources of features: one comes from the body keypoint detection branch as low-frequency global features, whereas the other is extracted from the hand area in the input image as high-frequency local features.
This feature composition utilizes body information for hand keypoint detection, and saves the computation of extracting high-level features for the hands, resulting in reduced runtime and improved accuracy.
%
\par
%
Further, we do not require a dataset where ground truth body, hands, and face reconstructions are all available at the same time: creating such data at sufficient variety is very difficult.
Instead, we only require existing part-specific datasets.
Our network features four task-specific modules that are trained individually with different types of data, while being end-to-end at inference.
The first module, \textit{DetNet}, takes a color image as input, estimates 3D body and hand keypoint coordinates, and detects the face location in the input image.
The second and third module, namely \textit{BodyIKNet} and \textit{HandIKNet}, take in body and hand keypoint positions and regress joint rotations along with shape parameters.
The last module, called \textit{FaceNet}, takes in a face image and predicts the shape, expression, albedo, and illumination parameters of the 3DMM face model \cite{tewari2017mofa}.
This modular network design enables us to jointly use the following data types:
\textit{1)} images with only body \textit{or} hand keypoint annotations;
\textit{2)} images with body \textit{and} hand keypoint annotations;
\textit{3)} images annotated with body joint angles;
\textit{4)} motion capture (MoCap) data with only body \textit{or} hand joint angles but without corresponding images;
and
\textit{5)} face images with 2D landmarks.
To train with so many data modalities, we propose an \textit{attention} mechanism to handle various data types in the same mini-batch during training, which guides the model to utilize the features selectively.
We also introduce a 2-stage body keypoint detection structure to cope with the keypoint discrepancy between different datasets.
The above multi-modal training enables our superior generalization across different benchmarks.
%
\par
%
Our contribution can be summarized as follows:
\setlist{nolistsep}
\begin{itemize}[noitemsep]
  \itemsep0em
  \item The first real-time approach that jointly captures 3D body, hands and face from a single color image.
  \item A novel network structure that combines local and global features and exploits inter-part correlations for hand keypoint detection, resulting in high computational efficiency and improved accuracy.
  \item The utilization of various data modalities supported by decoupled modules, an attention mechanism, and a 2-stage body keypoint detection structure, resulting in superior generalization.
\end{itemize}
%

\section{Related Work}
%
Human performance capture has a long research history.
Some methods are based on multi-view systems or a monocular depth camera to capture body~\cite{zhang20204d,joo2017panoptic}, hand~\cite{yuan2018depth,mueller2017real}, and face~\cite{ghosh2011multiview,roth2016adaptive}.
Although accurate, they are largely limited by the hardware requirements: multi-view systems are hard to setup while depth sensors do not work under bright sunlight.
This can be avoided by using a single RGB camera.
As our approach falls in the category of monocular methods, we focus on related works that only require a monocular image.
%
\par
%
\noindent \textbf{Body and Hand Capture.}
The very early researches~\cite{sigal2006measure,dantone2013human} propose to combine local features and spatial relationship between body parts for pose estimation.
With the advent of deep learning, new breakthrough is being made, from 2D keypoint detection~\cite{cao2018openpose, fang2017rmpe} to 3D keypoint estimation~\cite{tekin2016structured,habibie2019wild,mehta2017monocular,Artacho_2020_CVPR}.
In addition to sparse landmarks, recent approaches stress the task of producing a dense surface.
A series of statistical parametric models~\cite{anguelov2005scape,loper2015smpl,pavlakos2019expressive,joo2018total} are introduced and many approaches are proposed to estimate joint rotations for mesh animation.
Some of these work~\cite{mehta2017vnect,shimada2020physcap,xiang2019monocular} incorporate a separate inverse kinematics step to solve for joint rotations, while others~\cite{kanazawa2018end,kolotouros2019learning,habermann2020deepcap} regress model parameters from input directly.
To cope with the lack of detail in parametric models, some methods \cite{xu2018monoperfcap,habermann2019livecap,habermann2020deepcap} propose to use subject-specific mesh templates and perform dense tracking of the surface with non-rigid deformations.
Apart from model-based methods, model-free approaches also achieve impressive quality.
Various surface representations are proposed, including mesh~\cite{kolotouros2019convolutional}, per-pixel depth~\cite{gabeur2019moulding} and normal~\cite{smith2019facsimile}, voxels~\cite{zheng2019deephuman,jackson20183d}, and implicit surface functions~\cite{saito2019pifu,saito2020pifuhd}.
The research of hand capture has a similar history.
The task evolves from 2D keypoint detection~\cite{simon2017hand,wang2018mask}, to 3D keypoint estimation~\cite{zimmermann2017learning,mueller2018ganerated,doosti2020hope}, and finally dense surface recovery~\cite{boukhayma20193d,zhou2020monocular,zhang2019end,zhang2019interactionfusion} based on parametric models~\cite{romero2017embodied,tkach2016sphere}.
Methods that directly regresses mesh vertices are also proposed~\cite{moon2020deephandmesh,ge20193d,baek2020weakly}.
However, they all focus only on body or hands and failed to capture them jointly.
%
\par
%
\noindent \textbf{Face Capture.}
Early works~\cite{romdhani2005estimating,garrido2016reconstruction,thies2016face2face,wang2020emotion} reconstruct faces based on iterative optimization.
Deep learning approaches~\cite{richardson2017learning,tuan2017regressing} are also presented in the literature.
To cope with the problem of limited training data, semi- and self-supervised approaches are introduced~\cite{tewari2017mofa,tewari2018self,sengupta2018sfsnet,tewari2019fml}, where the models are trained in an analysis-by-synthesis fashion using differentiable rendering.
We refer to the surveys~\cite{zollhofer2018state,egger20203d} for more details.
%
\par
%
\noindent \textbf{Full Body Capture.}
Several recent works investigate the task of capturing body, face and hands simultaneously from a monocular color image.
The work of \cite{dope} estimates 3D keypoints of full body by distilling knowledge from part experts.
To obtain joint angles, previous works~\cite{xiang2019monocular,pavlakos2019expressive} propose a two-stage approach that first uses a network to extract keypoint information and then fits a body model onto the keypoints.
Choutas el al.~\cite{choutas2020monocular} regress model parameters directly from the input image and then apply hand/face-specific models to refine the capture iteratively.
Although they demonstrate promising results, they are all far from being real-time.
The shared shortcoming of their approaches is that they do not consider the correlation between body and hands.
In their work, body information is merely used to locate~\cite{xiang2019monocular,choutas2020monocular,pavlakos2019expressive} and initialize~\cite{choutas2020monocular} hands, while we argue that the high-level body features can help to deduce the hand pose~\cite{ng2020body2hands}.
Further, recent methods~\cite{xiang2019monocular,pavlakos2019expressive,choutas2020monocular} only capture facial expression, while our approach also recovers the facial identity in terms of geometry and color.
%

\section{Method}
%
\begin{figure*}[ht!]
  \centering
  \includegraphics[width=\textwidth]{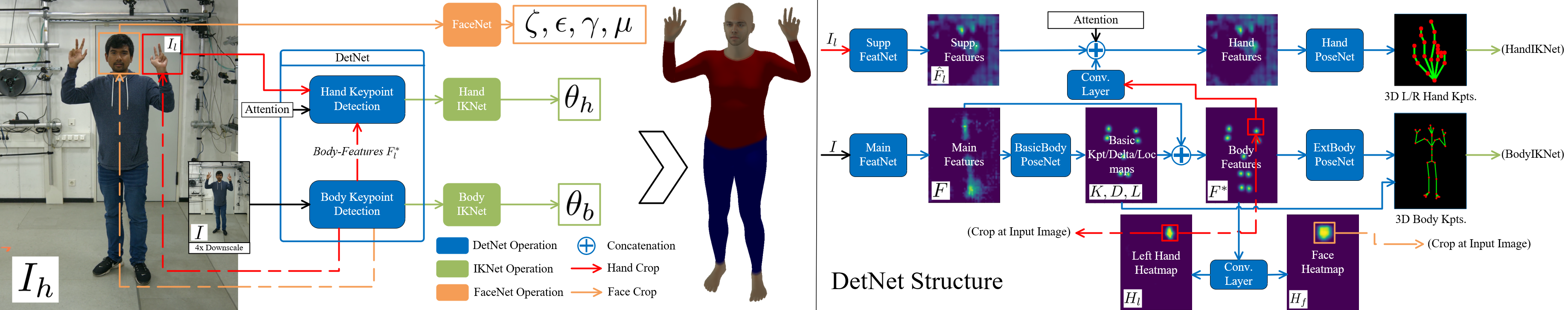}
  \caption{
    System overview and DetNet structure.
    Left: An input image $I_h$ is first downscaled by 4x for body keypoint detection and face/hand localization.
    The hand area is then cropped from $I_h$ to retrieve \textit{supp-features}, which are concatenated with processed \textit{body-features} for hand keypoint detection.
    Here, we use the attention channel to indicate the validity of \textit{body-features}.
    Body and hand 3D keypoint positions are fed into \textit{BodyIKNet} and \textit{HandIKNet} to estimate joint angles.
    The face area is cropped from $I_h$ and processed by \textit{FaceNet}.
    Finally, the parameters are combined to obtain a full mesh.
    Right: The detailed structure of \textit{DetNet}.
    Descriptions can be found in Sec.~\ref{sec:DetNet}.
    We only illustrate one hand for simplicity.
  }
  \label{fig:pipeline}
\end{figure*}
%
%
As shown in Fig.~\ref{fig:pipeline}, our method takes a color image as input, and outputs 2D and 3D keypoint positions, joint angles, and shape parameters of body and hands, together with facial expression, shape, albedo, and illumination parameters.
We then animate our new parametric model (Sec.~\ref{sec:mesh}) to recover a dense full body surface.
To leverage various data modalities, the whole network is trained as four individual modules:
\textit{DetNet} (Sec.~\ref{sec:DetNet}) that estimates body and hand keypoint positions from a body image, with our novel inter-part feature composition, the attention mechanism, and the 2-stage body keypoint detection structure;
\textit{BodyIKNet} and \textit{HandIKNet} (Sec.~\ref{sec:IKNet}) that estimate shape parameters and joint angles from keypoint coordinates for body and hands;
and \textit{FaceNet} (Sec.~\ref{sec:FaceNet}) that regresses face parameters from a face image crop.
%
%
\subsection{Full Body Model}
%
\label{sec:mesh}
\noindent \textbf{Body with Hands.}
We use the SMPLH-neutral \cite{romero2017embodied} model to represent the body and hands.
Specifically, SMPLH is formulated as
\begin{equation}
  T_B = \bar{T}_B + \beta E_\beta
\end{equation}
where $\bar{T}_B$ is the mean body shape with $N_B = 6890$ vertices, $E_\beta$ is the PCA basis accounting for different body shapes, and values in $\beta \in \mathbb{R}^{16}$ indicate PCA coefficients.
Given the body pose $\theta_b$ and the hand pose $\theta_h$, which represent the rotation of $J_B = 22$ body joints and $J_H = 15 \times 2$ hand joints, the posed mesh is defined as
\begin{equation}
  V_B = W(T_B, \mathcal{W}, \theta_b, \theta_{h})
\end{equation}
where $W(\cdot)$ is the linear blend skinning function and $\mathcal{W}$ are the skinning weights.
%
\par
%
\noindent \textbf{Face.}
For face capture, we adopt the 3DMM~\cite{blanz1999a} face model used in \cite{tewari2017mofa}.
Its geometry is given as
\begin{equation}
  V_{F} = \bar{V}_{F} + \zeta E_{\zeta} + \epsilon E_{\epsilon}
\end{equation}
where $\bar{V}_{F}$ is the mean face with $N_F = 53490$ vertices, $E_{\zeta}$ and $E_{\epsilon}$ are PCA bases that encode shape and expression variations, respectively.
$\zeta \in R^{80}$ and $\epsilon \in R^{64}$ are the shape and expression parameters to be estimated.
The face color is given by
\begin{equation}
  R = \bar{R} + \gamma E_{\gamma}
\end{equation}
\begin{equation}
  t_i = r_i \sum^{B^2}_{b=1}\mu_bH_b(n_i)
\end{equation}
where $R$ and $r_i$ are per vertex reflection, $\bar{R}$ is the mean skin reflectance, $E_{\gamma}$ is the PCA basis for reflectance, $t_i$ and $n_i$ are radiosity and normal of vertex $i$, and $H_b: \mathbb{R}^3 \to \mathbb{R}$ are the spherical harmonics basis functions.
We set $B^2 = 9$.
$\gamma \in R^{80}$ and $\mu \in \mathbb{R}^{3 \times 9}$ are albedo and illumination parameters.
%
\par
\begin{figure}[t]
  \centering
  \includegraphics[width=\linewidth]{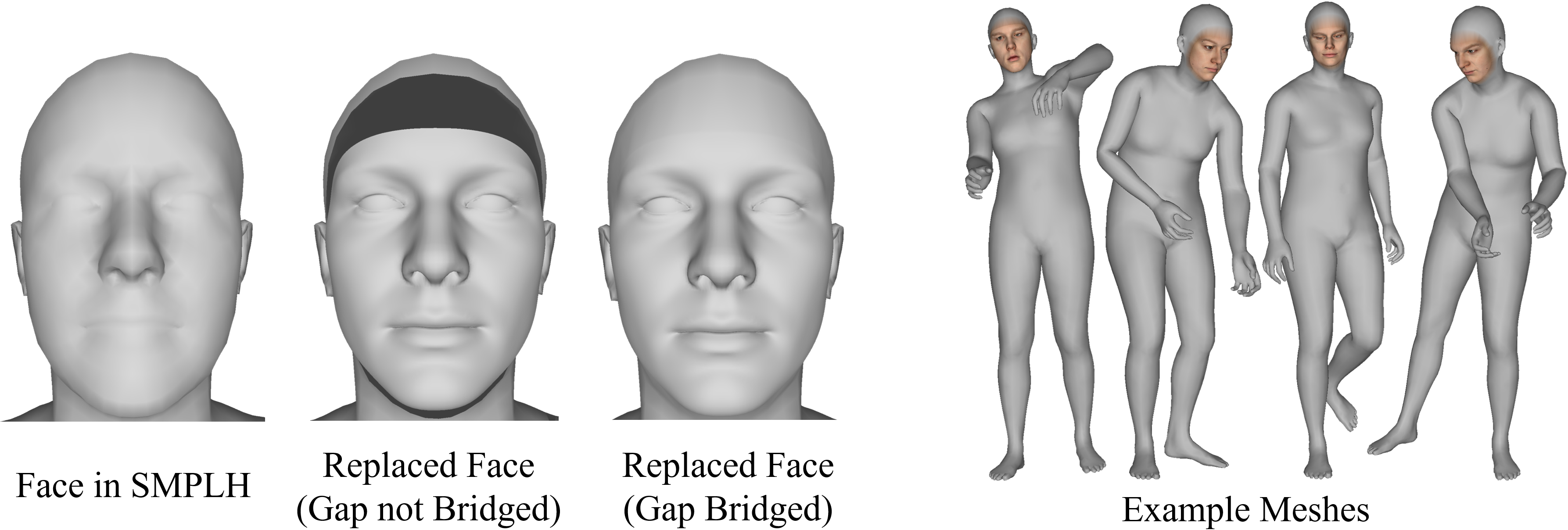}
  \caption{
    Our mesh model.
    From left to right:
    the original face in SMPLH;
    the replaced face (gap not bridged);
    the replaced face (gap bridged);
    example full body meshes.
  }
  \label{fig:mesh}
\end{figure}

%
\noindent \textbf{Combining Face and Body.}
To replace the SMPLH face with the 3DMM face, we manually annotate the face boundary $\mathcal{B}_b$ of SMPLH and the corresponding boundary $\mathcal{B}_f$ on the 3DMM face.
Then, a rigid transformation with a scale factor is manually set to align the face-excluded part of $\mathcal{B}_b$ and the face part of $\mathcal{B}_f$.
This manual work only needs to be performed once.
After bridging the two boundaries using Blender \cite{blender}, the face part rotates rigidly by the upper-neck joint using the head angles.
Unlike previous works~\cite{pavlakos2019expressive,joo2018total}, we do not simplify the face mesh.
Our model has more face vertices ($N'_F = 23817$) than the full body meshes of \cite{choutas2020monocular,pavlakos2019expressive} (10475 vertices) and \cite{joo2018total,xiang2019monocular} (18540 vertices), supports more expression parameters (64 versus 40~\cite{joo2018total,xiang2019monocular} and 10~\cite{choutas2020monocular,pavlakos2019expressive}), and embeds identity and color variation for face while others do not.
This design allows us to model face more accurately and account for the fact that humans are more sensitive to the face quality.
We show the combination process and full body meshes in Fig.~\ref{fig:mesh}.
%
%
%
\subsection{Keypoint Detection Network: DetNet}
%
\label{sec:DetNet}
The goal of our keypoint detection network, \textit{DetNet}, is to estimate 3D body and hand keypoint coordinates from the input image.
Particularly challenging is that body and hands have very different scales in an image so that a single network can barely deal with both tasks at the same time.
The naive solution would be to use two separate networks.
However, they would require much longer runtime, making real-time difficult to achieve.
Our key observation to solve this issue is that the high-level global features of the hand area extracted by the body keypoint estimation branch can be shared with the hand branch.
By combining them with the high-frequency local features additionally extracted from the hand area, expensive computation of hand high-level features is avoided, and body information for hand keypoint detection is provided, resulting in higher accuracy.
%
\par
%
\vspace{-3mm}
\subsubsection{Two-Stage Body Keypoint Detection\vspace{-1mm}}
It is a well-known issue that different body datasets have different sets of keypoint definitions, and the same keypoint is annotated differently in different datasets \cite{joo2018total}.
This inconsistency prevents the utilization of multiple datasets to improve the generalization ability.
To this end, instead of estimating all keypoints at once, we follow a two-stage manner for body keypoint detection.
We split the body keypoints into two subsets: \textit{basic body keypoints} which are shared by all body datasets without annotation discrepancy, and \textit{extended body keypoints} that are dataset-specific.
We use one \textit{BasicBody-PoseNet} to predict the \textit{basic body keypoints} for all datasets, and use different \textit{ExtBody-PoseNet}s to estimate the \textit{extended body keypoints} for different datasets.
This separation is essential for the multi-dataset training, and avoids \textit{BasicBody-PoseNet} to be biased to a specific dataset.
The \textit{-PoseNet} structure will be detailed in Sec.~\ref{sec:posenet}.
%
\par
%
The input of \textit{DetNet} is an image $I_h$ of resolution $768 \times 1024$ with one person as the main subject.
We bilinearly downscale it by a factor of 4 to get the low resolution image $I$, and feed it into the \textit{MainFeatNet}, a ResNet~\cite{he2016deep} alike feature extractor, to obtain main features $F$, which are fed into \textit{BasicBody-PoseNet} to estimate \textit{basic body keypoints}.
We then concatenate the features $F$ with the outputs of \textit{BasicBody-PoseNet} to get the body features $F^{*}$, which encodes high-level features and body information.
Finally, we use \textit{ExtBody-PoseNet} to predict the \textit{extended body keypoints} from $F^{*}$.
The \textit{basic body keypoints} and \textit{extended body keypoints} are combined to obtain the complete body keypoints.
%
\par
%
\vspace{-3mm}
\subsubsection{Hand Localization \vspace{-1mm}}
From the body features $F^*$, we use one convolutional layer to estimate left and right hand heat-maps $H_l$ and $H_r$.
For each hand, its heat-map $H$ is a one-channel 2D map where the value at each pixel represents the confidence that this pixel is occupied by the hand.
We use a sliding window to locate each hand from $H$, determined by its width $w$ and top-left corner location $(u, v)$, given by
\begin{equation}
  \arg \min_w: \max_{u, v} \sum_{i=u, j=v}^{i<u+w, j<v+w}h_{ij} > t * \sum_{i=0,j=0}^{i<a, j<b} h_{ij}
  \label{eq:hand_loc}
\end{equation}
where $h_{ij}$ is the confidence value of $H$ at pixel $(i, j)$; $a$ and $b$ are the width and height of $H$; and $t$ is a manually-set threshold value.
The intuition behind is to take the bounding box of minimal size that sufficiently contains the hand.
This heat-map based approach is consistent with the convolutional structure and the information of body embedded in $F^*$ is naturally leveraged in the estimation of $H$.
%
\par
%
\vspace{-3mm}
\subsubsection{Hand Keypoint Detection with Attention-based Feature Composition \vspace{-1mm}}
After hand localization, for the left and right hand, we crop $F^*$ at the area of the hands to get the corresponding features $F^*_l$ and $F^*_r$, referred to as \textit{body-features}.
They represent high-level global features.
Similarly, we crop the high resolution input image $I_h$ to get the left and right hand images $I_l$ and $I_r$, which are processed by \textit{SuppFeatNet} to obtain supplementary features $\hat{F}_l$ and $\hat{F}_r$, referred to as \textit{supp-features}.
They represent high-frequency local features.
For each hand, its corresponding \textit{body-features} are bilinearly resized and processed by one convolutional layer and then concatenated with its \textit{supp-features}.
The combined features are fed into \textit{Hand-PoseNet} to estimate hand keypoints.
This feature composition exploits the inter-part correlations between body and hands, and saves the computation of high-level features of the hand area by streaming directly from the body branch.
For time efficiency, \textit{SuppFeatNet} is designed to be a shallow network with only 8 ResNet blocks.
We use one \textit{SuppFeatNet} that handles $I_l$ and horizontally flipped $I_r$ at the same time.
The extracted features of $I_r$ are then flipped back.
On the other hand, we use two separate \textit{Hand-PoseNet}s for the two hands, as different hands focus on different channels of $F^*$.
%
\par
%
To leverage hand-only datasets for training, we further introduce an \textit{attention} mechanism that guides the hand branch to ignore \textit{body-features} when the body is not presented in the image.
Specifically, we additionally feed a one-channel binary-valued map into \textit{Hand-PoseNet} to indicate whether the \textit{body-features} are valid.
When the body is presented in the training sample, we set it to 1; otherwise, it is set to 0.
At inference, it is always set to 1.
%
\par
%
\vspace{-3mm}
\subsubsection{Face Localization \vspace{-1mm}}
\textit{DetNet} localizes the face in the input image using a face heat-map $H_f$ similarly as Eq.~\ref{eq:hand_loc}.
The face is cropped from the input image and later used to regress the face parameters by the separately trained \textit{FaceNet} module introduced in Sec.~\ref{sec:FaceNet}.
Different to the hands, \textit{FaceNet} only requires the face image and does not take $F^*$ as input.
This is based on our observation that the image input is sufficient for our fast \textit{FaceNet} to capture the face with high quality.
%
\par
%
\vspace{-3mm}
\subsubsection{Other Details \vspace{-1mm}}
\noindent \textbf{PoseNet Module.}
\label{sec:posenet}
The \textit{BasicBody-PoseNet}, the \textit{ExtBody-PoseNet}, and the \textit{Hand-PoseNet} share the same atomic network structure which comprises 6 convolutional layers to regress keypoint-maps $K$ (for 2D keypoint positions), delta-maps $D$ (for 3D bone directions), and location-maps $L$ (for 3D keypoint locations) from input features.
At inference, the coordinate of keypoint $i$ is retrieved from the location-map $L_i$ at the position of the maximum of the keypoint-map $K_i$.
The delta-map $D_i$ is for involving intermediate supervision.
Please refer to the supplementary document and ~\cite{mehta2017vnect} for more details.
The atomic loss function of this module is formulated as follows:
\begin{equation}
  \mathcal{L}_{p} = w_k \mathcal{L}_\mathrm{kmap} + w_d \mathcal{L}_\mathrm{dmap} + w_l \mathcal{L}_\mathrm{lmap}
\end{equation}
where
\begin{equation}
  \mathcal{L}_\mathrm{kmap} = || K^{\mathrm{GT}} - K ||^2_F
\end{equation}
\begin{equation}
  \mathcal{L}_\mathrm{dmap} = || K^{\mathrm{GT}} \odot (D^{\mathrm{GT}} - D) ||^2_F
\end{equation}
\begin{equation}
  \mathcal{L}_\mathrm{lmap} = || K^{\mathrm{GT}} \odot (L^{\mathrm{GT}} - L) ||^2_F \mathrm{.}
\end{equation}
$K$, $D$ and $L$ are keypoint-maps, delta-maps, and location-maps, respectively.
Superscript $\cdot^{\mathrm{GT}}$ denotes the ground truth, $||\cdot||_F$ is the Frobenius norm, and $\odot$ is the element-wise product.
$K^{\mathrm{GT}}$ is obtained by placing Gaussian kernels centered at the 2D keypoint locations.
$D^{\mathrm{GT}}$ and $L^{\mathrm{GT}}$ are constructed by tiling ground truth 3D keypoint coordinates and unit bone direction vectors to the size of $K^{GT}$.
$w_k$, $w_d$ and $w_l$ are hyperparameters to balance the terms.
For the training data without 3D labels, we set $w_d$ and $w_l$ to 0.
%
\par
%
\noindent \textbf{Full Loss.}
The full loss function of the \textit{DetNet} is defined as
\begin{equation}
  \lambda_b \mathcal{L}_{p}^{b} + \lambda_h (\mathcal{L}_{p}^{lh} + \mathcal{L}_{p}^{rh} + \mathcal{L}_{h}) + \lambda_f \mathcal{L}_{f} \mathrm{.}
\end{equation}
$\mathcal{L}_{p}^{b}$, $\mathcal{L}_{p}^{lh}$, and $\mathcal{L}_{p}^{rh}$ are the keypoint detection losses for body, left hand and right hand, respectively.
\begin{equation}
  \mathcal{L}_h = || H^{\mathrm{GT}}_{l} - H_l ||^2 + || H^{\mathrm{GT}}_{r} - H_r ||^2
\end{equation}
supervises hand heat-maps for hand localization.
Similarly,
\begin{equation}
  \mathcal{L}_f = || H^{\mathrm{GT}}_{f} - H_f ||^2
\end{equation}
supervises the face heat-map.
$H_f^{\mathrm{GT}}$, $H_{l}^{\mathrm{GT}}$, and $H_{r}^{\mathrm{GT}}$ are constructed by taking the maximum along the channel axis of the keypoint-maps to obtain a one-channel confidence map.
$\lambda_b$, $\lambda_h$, and $\lambda_f$ are hyperparameters which are set to 0 when the corresponding parts are not in the training sample.
%
\par
%
\noindent \textbf{Global Translation.}
All monocular approaches suffer from depth-scale ambiguity.
In \textit{DetNet}, the estimated keypoint positions are relative to the root keypoint.
However, when the camera intrinsics matrix $C$ and the length of any bone $l_{cp}$ are known, the global translation can be determined based on
\begin{equation}
  l_{cp} = || C^{-1} z_p\begin{bmatrix} u_p \\ v_p \\ 1 \end{bmatrix} -
  C^{-1} (z_p + d_c - d_p) \begin{bmatrix} u_w \\ v_w \\ 1 \end{bmatrix} ||_2 \mathrm{.}
  \label{eq:glb_trans}
\end{equation}
Here, the subscript $\cdot_c$ and $\cdot_p$ denote the child and parent keypoint of bone $l_{cp}$;
$u$ and $v$ are 2D keypoint positions;
$d$ refers to the root-relative depth;
and $z_p$ is the absolute depth of keypoint $p$ relative to the camera.
In Eq.~\ref{eq:glb_trans}, $z_p$ is the only unknown variable that can be solved in closed form.
When $z_p$ is known, the global translation can be computed with the camera projection formula.
%

\subsection{Inverse Kinematics Network: IKNet}
%
\label{sec:IKNet}
Sparse 3D keypoint positions are not sufficient to drive CG character models.
To animate mesh models and obtain dense surface, joint angles need to be estimated from sparse keypoints.
This task is known as inverse kinematics (IK).
Typically, the IK task is tackled with iterative optimization methods~\cite{bogo2016keep,guan2009estimating,xiang2019monocular,xu2018monoperfcap,habermann2019livecap,tkach2016sphere}, which are sensitive to initialization, take longer time, and need hand-crafted priors.
Instead, we use a fully connected neural network module, referred to as \textit{IKNet}, to regress joint angles from keypoint coordinates, similar to~\cite{zhou2020monocular}.
Trained with additional MoCap data, \textit{IKNet} learns a pose prior implicitly from the data, and as a result further decreases keypoint position errors.
Due to the end-to-end architecture, \textit{IKNet} achieves superior runtime performance, which is crucial for being real-time.
%
\par
%
In particular, \textit{IKNet} is a fully connected network that takes in keypoint coordinates and outputs joint rotations $\theta_b$ and $\theta_h$ for body and hands.
The main difference between our approach and~\cite{zhou2020monocular} is that we use relative 6D rotation~\cite{zhou2019continuity} as the output formulation, and our network additionally estimates the shape parameters $\beta$ and a scale factor $\alpha$.
Since there is little MoCap data that contains body and hand joint rotations simultaneously, and synthesizing such data is not guaranteed to be anatomically correct, we train \textit{BodyIKNet} and \textit{HandIKNet} to estimate $\theta_b$ and $\theta_h$ separately, instead of training a single network that regresses all joint angles.
The loss terms are defined as:
\begin{equation}
  \lambda_{\alpha} L_{\alpha} + \lambda_{\beta} L_{\beta} + \lambda_{\theta} L_{\theta} + \lambda_{\chi} L_{\chi} + \lambda_{\bar{\chi}} L_{\bar{\chi}} \mathrm{.}
\end{equation}
Here, $L_{\alpha}$, $L_{\beta}$, $L_{\theta}$, $L_{\chi}$, and $L_{\bar{\chi}}$ are L2 losses for the scale factor $\alpha$, shape parameters $\beta$, joint rotations $\theta$, keypoint coordinates after posing $\chi$, and keypoint coordinates at the reference pose $\bar{\chi}$.
$\lambda_{\cdot}$ are the weights for different terms.
%

\subsection{Face Parameters Estimation: FaceNet}
%
\label{sec:FaceNet}
We adopt a convolutional module, named \textit{FaceNet}, to estimate shape, expression, albedo and illumination parameters of a statistical 3DMM face model~\cite{blanz1999a} from a face-centered image.
The face image is obtained by cropping the original high-resolution image according to the face heat-map estimated by \textit{DetNet}.
Compared with previous full body capture works~\cite{xiang2019monocular,pavlakos2019expressive,joo2018total,choutas2020monocular} that only estimate facial expression, our regression of shape, albedo and illumination gives more personalized and realistic results.
\textit{FaceNet} is originally proposed and pre-trained by Tewari et al. \cite{tewari2017mofa}.
As the original model in~\cite{tewari2017mofa} is sensitive to the size and location of the face in the image, we finetune it with the face crops produced by the \textit{DetNet} for better generalization.
%

\section{Experiments}
%
\subsection{Datasets and Evaluation Metrics}
%
The following datasets are used to train \textit{DetNet}:
\textit{1)} body-only datasets: HUMBI~\cite{yu2020humbi}, MPII3D~\cite{mehta2017monocular}, HM36M~\cite{ionescu2013human3}, SPIN~\cite{kolotouros2019learning}, MPII2D~\cite{andriluka20142d}, and COCO~\cite{lin2014microsoft};
\textit{2)} hand-only datasets: FreiHand~\cite{zimmermann2019freihand}, STB~\cite{zhang2017a}, and CMU-Hand~\cite{simon2017hand};
\textit{3)} body with hands dataset: MTC~\cite{joo2018total}.
Here, MPII2D, COCO, and CMU-Hand only have 2D labels, but they are helpful for generalization since they are in-the-wild.
Please refer to the supplementary document for more details on these datasets.
We utilize AMASS~\cite{mahmood2019amass}, HUMBI and SPIN to train \textit{BodyIKNet}, and use the MoCap data from MANO~\cite{romero2017embodied} to train \textit{HandIKNet} following the method of \cite{zhou2020monocular}.
The training data for \textit{HandIKNet} and \textit{BodyIKNet} are augmented as in \cite{zhou2020monocular}.
\textit{FaceNet} is pre-trained on the VoxCeleb2~\cite{Chung18b} dataset following~\cite{tewari2017mofa}, and fine-tuned with face images from MTC.
%
\par
%
We evaluate body predictions on MTC, HM36M, MPII3D, and HUMBI, using the same protocol as in~\cite{xiang2019monocular} (MTC, HM36M) and~\cite{mehta2017vnect} (MPII3D).
On HUMBI, we select 15 keypoints for evaluation to be consistent with other datasets, and ignore the keypoints outside the image.
For hand evaluation we use MTC and FreiHand.
Since not all the test images in MTC have both hands annotated, we only evaluate on the samples where both hands are labeled, referred to as MTC-Hand.
We use Mean Per Joint Position Error (MPJPE) in millimeter (mm) as the metric for body and hand pose estimation, and follow the convention of previous works to report results without (default) and with (indicated by $^\ddagger$ and ``PA'') rigid alignment by performing Procrustes analysis.
As \cite{choutas2020monocular} outputs the SMPL mesh, we use a keypoint regressor to obtain HM36M-style keypoint predictions, similar to~\cite{kolotouros2019learning,kanazawa2018end}.
We evaluate \textit{FaceNet} on the face images cropped from MTC test set by using 2D landmark error and per channel photometric error as the metric.
We use PnP-RANSAC~\cite{fischler1981random} and PA alignment to estimate camera pose for projection and error computation of the face.
%

\subsection{Qualitative Results}
\begin{figure}[t]
  \centering
  \includegraphics[width=\linewidth]{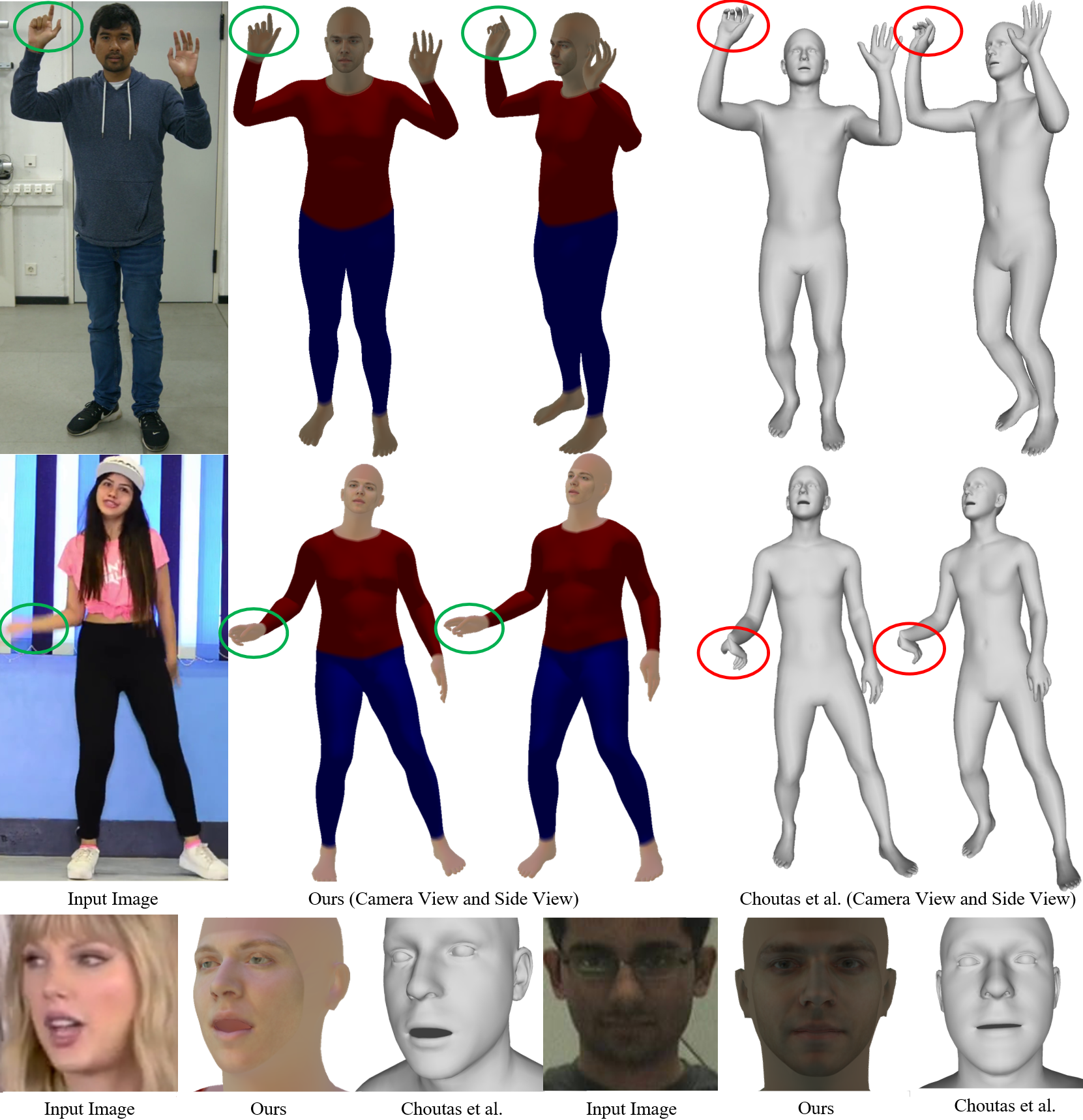}
  \caption{
    Qualitative results.
    From top to bottom:
    \textit{1)} our method captures subtle gestures while \cite{choutas2020monocular} is over-smooth;
    \textit{2)} our hand pose is consistent with the wrist and arm while \cite{choutas2020monocular} is anatomically incorrect;
    \textit{3)} our faces are more personalized and realistic due to the variation in identity-dependent facial geometry and albedo.
  }
  \label{fig:quat_cmp}
\end{figure}

\begin{figure}[t]
  \centering
  \includegraphics[width=\linewidth]{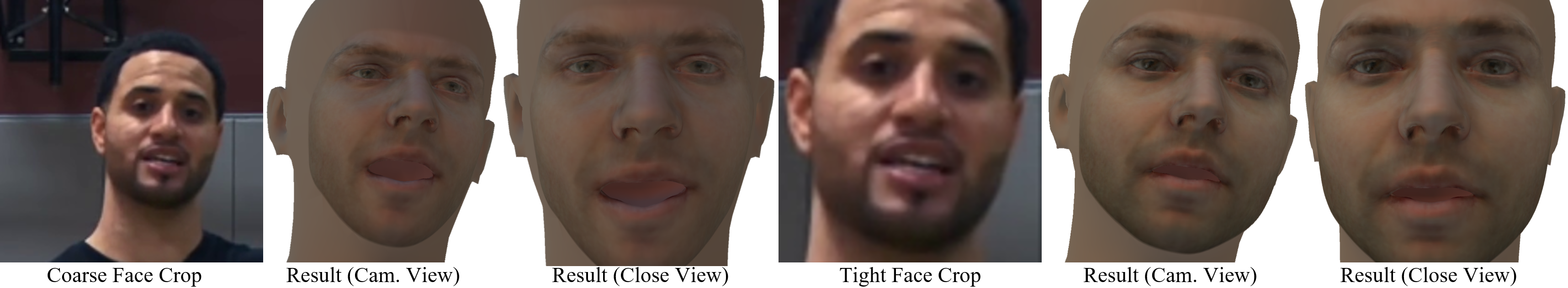}
  \caption{
    Comparison on face crop.
    A coarse face crop is already sufficient for face capture, while a tighter one further improves quality.
  }
  \label{fig:face_crop}
\end{figure}

%
We present qualitative results in Fig.~\ref{fig:quat_cmp} and compare with the state-of-the-art approach of Choutas et al. \cite{choutas2020monocular}.
Despite much faster inference speed, our model gives results with equal visual quality.
In the first row we show that our model captures detailed hand poses while \cite{choutas2020monocular} gives over-smooth estimation.
This is because of our utilization of high-frequency local features extracted from the high-resolution hand image.
In the second row, we demonstrate that our hand pose is consistent with the wrist and arm, while the result of \cite{choutas2020monocular} is anatomically incorrect.
This is due to our utilization of body information for hand pose estimation.
We demonstrate in the third row that with variations in facial shape and color, our approach provides highly personalized capture results, while \cite{choutas2020monocular} lacks identity information.
In Fig.~\ref{fig:face_crop} we compare the face capture results of coarse and tight face crops.
The result on the loosely cropped image already captures the subject very well (left), and a tighter bounding box obtained from a third party face detector \cite{king2009dlib} based on the coarse crop further improves the quality (right).
Unless specified, the presented results in the paper are all based on tight face crops.
As our approach does not estimate camera pose, for overlay visualization, we adopt PnP-RANSAC~\cite{fischler1981random} and PA alignment to align our 3D and 2D predictions.
The transformations are rigid and no information of ground truth is used.
Please refer to the supplemental material for more results.
%
%
%
%
%
%
\begin{figure}[t]
  \centering
  \includegraphics[width=\linewidth]{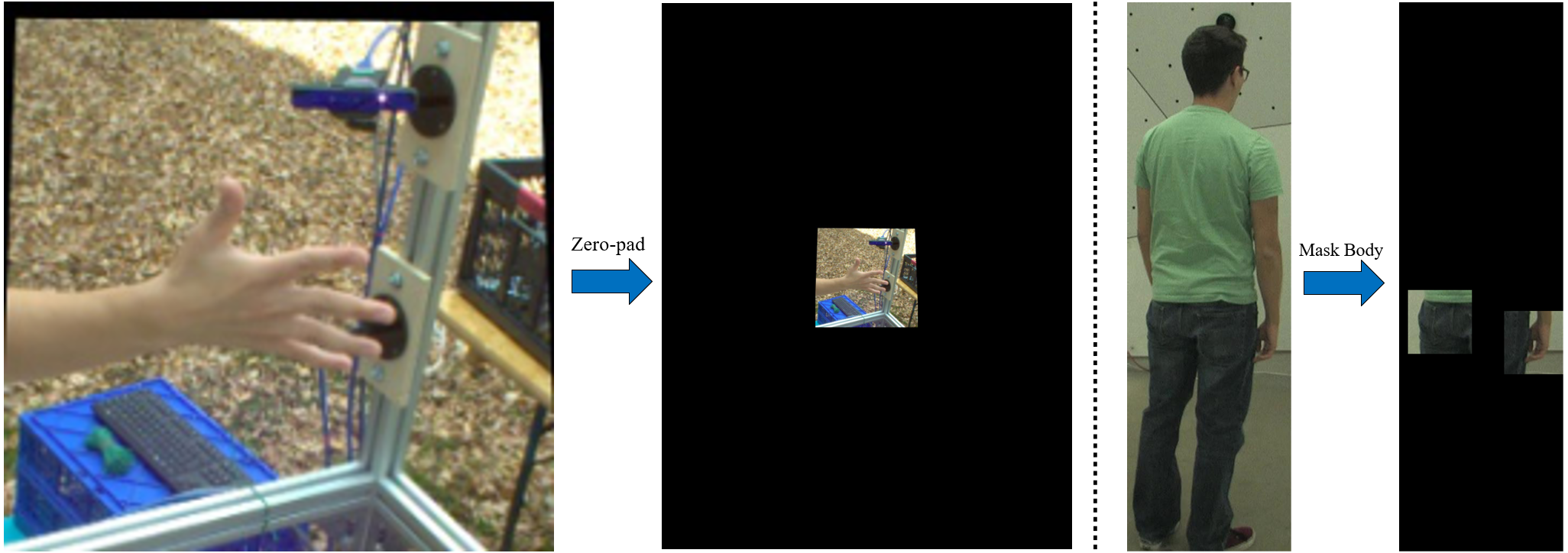}
  \caption{
    Samples from test data.
    Left: we zero-pad the hand-only image from FreiHand to evaluate our model, which is disadvantageous for us.
    Right: we mask the body and only keep the hand regions visible to construct the MTC-Hand-Mask test set.
  }
  \label{fig:test_img}
\end{figure}
%

\subsection{Quantitative Results}
%
\noindent \textbf{Runtime.}
Runtime performance is crucial for a variety of applications, thus real-time capability is one of our main goals.
In Tab.~\ref{tab:runtime}, we report the runtime of each subtask in milliseconds (ms) on a commodity PC with an Intel Core i9-10920X CPU and an Nvidia 2080Ti GPU.
We use -B and -H to indicate body and hand sub-tasks.
Due to the efficient inter-part feature composition, it takes only 10.3ms to estimate keypoint positions of two hands, which is two times faster than the lightweight method of \cite{zhou2020monocular}.
The end-to-end \textit{IKNet} takes 2.68ms in total, which is nearly impossible for traditional iterative optimization-based IK solvers.
The optional face detector \cite{king2009dlib} takes 7ms, without breaking the real-time limitation (25.5fps).
%
%
\begin{table}[t]
  \centering
  \begin{subtable}[c]{\linewidth}
  \centering
  \resizebox{\linewidth}{!}{
    \begin{tabular}{ |c|c|c|c|c|c|c|c| }
    \hline
    Module & DetNet-B & DetNet-H & IKNet-B & IKNet-H & FaceNet & Total \\
    \hline
    Runtime & 16.9 & 10.3 & 1.51 & 1.17 & 1.92 & 32.1 \\
    \hline
    \end{tabular}
  }
  \end{subtable}
  \begin{subtable}[c]{\linewidth}
  \centering
  \resizebox{\linewidth}{!}{
    \begin{tabular}{ |c|c|c|c|c|c| }
    \hline
    Method & \textbf{Ours} & Kanazawa~\cite{kanazawa2018end} & Choutas~\cite{choutas2020monocular} & Xiang~\cite{xiang2019monocular} & Pavlakos~\cite{pavlakos2019expressive} \\
    \hline
    Runtime & \textbf{32.1} & 60 & 160 & 20000 & $\sim$50000 \\
    \hline
    FPS & \textbf{31.1} & 16.7 & 6.25 & 0.05 & $\sim$0.02 \\
    \hline
    \end{tabular}
  }
  \end{subtable}
  \caption{
    Runtime analysis in milliseconds and frames per second (FPS).
    Top: runtime of each subtask in our method.
    Bottom: comparison with previous works.
  }
  \label{tab:runtime}
\end{table}
%
\par
%
\noindent \textbf{Body Pose Estimation.}
In Tab.~\ref{tab:body_quant_comp}, we report quantitative evaluation for body keypoint detection of \textit{DetNet}, and compare with other state-of-the-art approaches.
Despite \textit{DetNet} is extremely fast, it is still comparable with the top models in terms of accuracy.
We also evaluate previous works on HUMBI although they were not trained on the train split.
Notably, their accuracy significantly drops as their generalization across datasets is limited.
In contrast, our approach performs similarly well across all datasets due to the multi-dataset training, indicating a better generalization ability.
In Tab.~\ref{tab:body_quant_humbi}, we compare the results after \textit{BodyIKNet} on HUMBI with different sources of shape parameters: IK-$\beta$ uses the shape parameters estimated by \textit{BodyIKNet}, and GT-$\beta$ uses the ground truth shape parameters.
Due to the additional knowledge of the pose prior learned from MoCap data, \textit{BodyIKNet} decreases the keypoint error.
After PA alignment, the error of IK-$\beta$ is very close to GT-$\beta$, indicating that the body shape estimation is also accurate.
%
%
\begin{table}[t]
  \centering
  \resizebox{\linewidth}{!}{
    \begin{tabular}{ |c|c|c|c|c| }
    \hline
    \multirow{2}{*}{Method} & \multicolumn{4}{c|}{MPJPE (mm)} \\
    \cline{2-5} & HM36M & MPII3D & MTC &  HUMBI \\
    \hline
    Xiang et al.~\cite{xiang2019monocular} & \textbf{58.3} & - & \textbf{63.0} & - \\
    \hline
    Kolotouros et al.~\cite{kolotouros2019learning} & \textbf{41.1}$^\ddagger$ & \textbf{105.2} & - & 101.7$^{\ddagger \mathsection}$ \\
    \hline
    Choutas et al.~\cite{choutas2020monocular} & 54.3$^\ddagger$ & - & - & 67.2$^{\ddagger \mathsection}$ \\
    \hline
    Kanazawa et al.~\cite{kanazawa2018end} & 56.8$^\ddagger$ & 124.2 & - & 84.2$^{\ddagger \mathsection}$ \\
    \hline
    \textbf{DetNet} & 64.8 & 116.4 & 66.8 & \textbf{43.5} \\
    \hline
    \textbf{DetNet} (PA) & 50.3$^\ddagger$ & \textbf{77.0}$^\ddagger$ & \textbf{61.5}$^\ddagger$ & \textbf{32.5}$^\ddagger$ \\
    \hline
    \end{tabular}
  }
  \caption{
    Body MPJPE on public datasets.
    Our model has competitive results across all datasets while being much faster.
    $^\mathsection$ means the model is not trained on the train split.
  }
  \label{tab:body_quant_comp}
\end{table}
%
%
\begin{table}[t]
  \centering
  \resizebox{\linewidth}{!}{
    \begin{tabular}{ |c|c|c|c| }
    \hline
    Metric & DetNet & \textbf{DetNet+IKNet (IK-$\pmb{\beta}$)} & DetNet+IKNet (GT-$\beta$)  \\
    \hline
    MPJPE & 43.5 & 43.3 & \textbf{39.9} \\
    \hline
    MPJPE (PA) & 32.5$^\ddagger$ & 31.6$^\ddagger$ & \textbf{31.2}$^\ddagger$ \\
    \hline
    \end{tabular}
  }
  \caption{
    Body MPJPE on HUMBI.
    We demonstrate that incorporating \textit{BodyIKNet} further lowers error.
    The small gap between IK-$\beta$ and GT-$\beta$ indicates the high accuracy of body shape estimation.
  }
  \label{tab:body_quant_humbi}
\end{table}
%
\par
%
\noindent \textbf{Hand Pose Estimation.}
We report our results for hand pose estimation in Tab.~\ref{tab:hand_quant_comp}.
The results after IK are based on the shape parameters estimated by \textit{HandIKNet}.
On the MTC-Hand test set, our mean error is only 9.3mm.
We attribute the 1.1mm increase of error after IK to the difference in keypoint definitions between our hand model (SMPLH) and the MTC hand model, as the bone length difference is 25\% on average.
When it comes to FreiHand, our error increases.
This is because FreiHand is a hand-only dataset, while in our method hand pose deeply relies on body information.
Since we do not have a hand-specific module, to evaluate on FreiHand, we have to zero-pad the hand image to the full size and feed it into the model (Fig.~\ref{fig:test_img}) as if body is presented.
Despite this non-ideal setup, after IK, our error is still comparable to \cite{choutas2020monocular}, and outperforms \cite{zhou2020monocular}
which is not trained on FreiHand.
Note that the previous methods in Tab.~\ref{tab:hand_quant_comp} are not trained on the train split of MTC and cannot compare with us directly on MTC-Hand.
%
%
\begin{table}[t]
  \centering
  \resizebox{\linewidth}{!}{
    \begin{tabular}{ |c|c|c|c| }
    \hline
    \multirow{2}{*}{Method} & \multicolumn{3}{c|}{MPJPE (mm)} \\
    \cline{2-4} & MTC-Hand (left) & MTC-Hand (right) & FreiHand \\
    \hline
    Choutas et al.~\cite{choutas2020monocular} & 13.0$^{\ddagger\mathsection}$ & 12.2$^{\ddagger\mathsection}$ & \textbf{12.2}$^\ddagger$ \\
    \hline
    Zhou et al.~\cite{zhou2020monocular} & 16.1$^{\ddagger\mathsection}$ & 15.6$^{\ddagger\mathsection}$ & 21.8$^{\ddagger\mathsection}$ \\
    \hline
    \textbf{DetNet} & \textbf{15.1} & \textbf{13.8} & - \\
    \hline
    \textbf{DetNet} (PA) & \textbf{8.50}$^\ddagger$  & \textbf{7.90}$^\ddagger$ & 24.2$^\ddagger$ \\
    \hline
    \textbf{DetNet + IKNet} (PA) & 9.42$^\ddagger$ & 9.10$^\ddagger$ & 15.7$^\ddagger$ \\
    \hline
    \end{tabular}
  }
  \caption{
    Hand MPJPE on public datasets.
    Our model has the lowest error on MTC-Hand where the body information is available, and is comparable on FreiHand even the body is absent.
    $^\mathsection$ means the model is not trained on the train split.
  }
  \label{tab:hand_quant_comp}
\end{table}
%
\par
%
\noindent \textbf{Face Capture.}
In Tab.~\ref{tab:face_quant_comp}, we evaluate \textit{FaceNet} on the face crops from the MTC test set (MTC-Face).
Compared with typical datasets, the faces in MTC-Face are more blurry and challenging.
Our \textit{FaceNet} gives better results than \cite{tewari2017mofa} on such in-the-wild samples, and a tighter face bounding box (denoted by postfix ``T'') further lowers error.
Please refer to the supplementary document for more evaluation on face.
\begin{table}[t]
  \centering
  \resizebox{\linewidth}{!}{
    \begin{tabular}{ |c|c|c|c| }
    \hline
    Metric & Tewari et al.~\cite{tewari2017mofa} & \textbf{FaceNet} & \textbf{FaceNet-T} \\
    \hline
    Landmark Err. & 4.70 & 3.43 & \textbf{3.37} \\
    \hline
    Photometric Err. & 0.0661 & 0.0447 & \textbf{0.0444} \\
    \hline
    \end{tabular}
  }
  \caption{
    Landmark error in pixel and photometric error per channel on MTC-Face.
    \textit{FaceNet} performs better than \cite{tewari2017mofa} on these challenging samples, and a tighter bounding box further improves accuracy.
  }
  \label{tab:face_quant_comp}
\end{table}
%

\subsection{Ablation Study}
%
\label{sec:ablation}
\noindent \textbf{Feature Composition.}
The inter-part feature composition from body to hands is critical to reduce runtime and improve hand pose accuracy.
To examine this design, we train the following models for comparison:
\textit{1)} \textit{DetNet-S}(upplementary) where the hand branch estimates hand pose only from \textit{supp-features} $\hat{F}$ and does not take any information from body except hand localization;
\textit{2)} \textit{DetNet-B}(ody) where the hand branch estimates hand pose only from \textit{body-features} $F^*$ and does not see the high-resolution input image.
To further examine the importance of body information for hand keypoint detection, we additionally construct a test set derived from MTC-Hand, called MTC-Hand-Mask, where the body area is masked and only the hands are visible (Fig.~\ref{fig:test_img}).
The results are reported in Tab.~\ref{tab:ablation_hand}.
On MTC-Hand, because of the utilization of body information, the error of \textit{DetNet} is lower than \textit{DetNet-S} by 28\%.
When it comes to FreiHand and MTC-Hand-Mask, the gap between \textit{DetNet} and \textit{DetNet-S} shrinks to 4\% and \mbox{-5\%}.
This is due to the missing body information in these two test sets, which indicates that the \textit{body-features} indeed contribute to the hand keypoint detection.
\textit{DetNet-B} always performs worse than \textit{DetNet}.
This is because \textit{body-features} are extracted from the low-resolution image where the hands are too blurry and cover only a few pixels.
This comparison indicates the importance of \textit{supp-features}.
%
%
\begin{table}[t]
  \centering
  \resizebox{\linewidth}{!}{
    \begin{tabular}{ |c|c|c|c| }
    \hline
    \multirow{2}{*}{Method} & \multicolumn{3}{c|}{MPJPE (mm)} \\
    \cline{2-4}  & MTC-Hand & MTC-Hand-Mask & FreiHand \\
    \hline
    DetNet-S(upplementary) & 18.4 & 31.7 & \textbf{23.1}$^\ddagger$ \\
    \hline
    DetNet-B(ody) & 17.2 & 37.5 & 26.8$^\ddagger$ \\
    \hline
    \textbf{DetNet} & \textbf{14.4} & \textbf{30.6} & 24.2$^\ddagger$ \\
    \hline
    \end{tabular}
  }
  \caption{
    Ablation study on \textit{body-features} and \textit{supp-features}.
    The comparison between the three versions demonstrates the help of $F^*$ and $\hat{F}$ in the hand pose estimation task.
  }
	\label{tab:ablation_hand}
\end{table}
%
\par
%
\noindent \textbf{Data Modalities.}
The advantage of using MoCap data is examined in Tab.~\ref{tab:body_quant_humbi} where \textit{IKNet} lowers the error.
To evaluate the attention mechanism and multiple image datasets, we train the following models:
\textit{1)} \textit{DetNet-U}(niform) which is trained without the attention mechanism, i.e. we treat hand-only data as if body is presented by always setting the attention channel to 1;
\textit{2)} \textit{DetNet-O}(verfitted) which is trained on the only dataset where body and hands are annotated simultaneously, namely MTC;
\textit{3)} \textit{DetNet-I}(ndoor) that only uses the training data with 3D annotations (usually indoor) without any 2D-labeled data (usually in-the-wild).
To account for different keypoint definitions, we only evaluate \textit{basic body keypoints}, except for MTC where all the models are trained on.
As shown in Tab.~\ref{tab:ablation_body}, \textit{DetNet-U} generally performs worse than \textit{DetNet}, indicating that the attention mechanism helps during training.
\textit{DetNet-O} has poor cross-dataset generalization and only performs well on MTC-Hand.
This illustrates the importance of the multi-dataset training strategy, which is enabled by our 2-stage keypoint detection structure.
Finally, the inferior of \textit{DetNet-I} to \textit{DetNet} demonstrates the help of in-the-wild images, although they only have 2D annotations.
Please refer to the supplementary video for more evaluation on the training data.
%
%
\begin{table}[t]
  \resizebox{\linewidth}{!}{
    \begin{tabular}{ |c|c|c|c|c|c| }
    \hline
    \multirow{2}{*}{Method} & \multicolumn{5}{c|}{MPJPE (mm)} \\
    \cline{2-6} & HM36M & MPII3D & MTC & HUMBI & MTC-Hand \\
    \hline
    DetNet-U(niform) & 57.9$^\ddagger$ & 99.9$^\ddagger$ & \textbf{64.6} & 59.1 & 14.7 \\
    \hline
    DetNet-O(verfitted) & 272.2$^\ddagger$ & 297.9$^\ddagger$ & 67.7 & 289.4 & \textbf{13.8} \\
    \hline
    DetNet-I(ndoor) & 61.7$^\ddagger$ & 95.7$^\ddagger$ & 64.8 & 63.1 & 15.1 \\
    \hline
    \textbf{DetNet} & \textbf{57.5}$^\ddagger$ & \textbf{90.1}$^\ddagger$ & 66.8 & \textbf{52.5} & 14.4 \\
    \hline
    \end{tabular}
  }
  \caption{
    Ablation study on training data.
    The gap between \textit{DetNet-U} and \textit{DetNet} shows the help of the attention mechanism.
    \textit{DetNet-O} and \textit{DetNet-I} only perform well on a few datasets, while \textit{DetNet} has the best cross-dataset accuracy.
  }
	\label{tab:ablation_body}
\end{table}

%

\section{Conclusion}
%
We present the first real-time approach to capture body, hands, and face from an RGB image.
The accuracy and time efficiency comes from our network design that exploits inter-part relationship between body and hands.
By training the network as separate modules, we leverage multiple data sources and achieve superior generalization.
Further, our approach captures personalized face with both expression and identity-dependent shape and albedo.
Future directions can involve temporal information for smoother results.
%

{
  \small
  \bibliographystyle{ieee_fullname}
  \bibliography{reference}
}

\end{document}